\documentclass{article}

\usepackage{PRIMEarxiv}

\usepackage[utf8]{inputenc} 
\usepackage[T1]{fontenc}    
\usepackage[hidelinks]{hyperref}       
\usepackage{url}            
\usepackage{booktabs}       
\usepackage{amsfonts}       
\usepackage{nicefrac}       
\usepackage{microtype}      
\usepackage{lipsum}
\usepackage{fancyhdr}       
\usepackage{graphicx}       
\graphicspath{{media/}}     

\usepackage[round]{natbib}
\usepackage{algorithm}

\title{AlphaD3M: Machine Learning Pipeline Synthesis}

\author{%
\begin{minipage}[t]{12cm}
\normalsize \textbf{\textsc{Iddo Drori}} \hfill
\normalsize \href{mailto:idrori@nyu.edu}{idrori@nyu.edu} \\
\textbf{\textsc{Yamuna Krishnamurthy}} \hfill
\normalsize \href{mailto:yamuna@nyu.edu}{yamuna@nyu.edu} \\
\textbf{\textsc{Remi Rampin}} \hfill
\normalsize \href{mailto:remi.rampin@nyu.edu}{remi.rampin@nyu.edu} \\
\textbf{\textsc{Raoni de Paula Lourenco}} \hfill
\normalsize \href{mailto:raoni@nyu.edu}{raoni@nyu.edu} \\
\textbf{\textsc{Jorge Piazentin Ono}} \hfill
\normalsize \href{mailto:jorgehpo@nyu.edu}{jorgehpo@nyu.edu} \\
\textbf{\textsc{Kyunghyun Cho}} \hfill
\normalsize \href{mailto:kyunghyun.cho@nyu.edu}{kyunghyun.cho@nyu.edu} \\
\textbf{\textsc{Claudio Silva}} \hfill
\normalsize \href{mailto:csilva@nyu.edu}{csilva@nyu.edu} \\
\textbf{\textsc{Juliana Freire}} \hfill
\normalsize \href{mailto:juliana.freire@nyu.edu}{juliana.freire@nyu.edu}
\end{minipage}
}

\begin{document}
\bibliographystyle{abbrvnat}  
\maketitle

\begin{abstract}
We introduce AlphaD3M, an automatic machine learning (AutoML) system based on meta reinforcement learning using sequence models with self play. AlphaD3M is based on edit operations performed over machine learning pipeline primitives providing explainability. We compare AlphaD3M with state-of-the-art AutoML systems: Autosklearn, Autostacker, and TPOT, on OpenML datasets. AlphaD3M achieves competitive performance while being an order of magnitude faster, reducing computation time from hours to minutes, and is explainable by design.
\end{abstract}

\section{Introduction}
\label{introduction}
Automatic machine learning (AutoML) aims to learn how to learn. Given a dataset, a well defined task, and performance criteria, the goal is to solve the task with respect to the dataset while optimizing performance.
Existing systems have focused on a relatively small set of machine learning primitives, with a few tasks \citep{feurer2015autosklearn}, or on a small set of datasets \citep{chen2018autostacker}, or on numerous datasets within specific domains \citep{olson2016tpot}. 

DARPA's Data Driven Discovery of Models (D3M) program pushes this vision further and proposes to develop infrastructure to automate model discovery, i.e., solve any task on any dataset specified by the user. Using a broad set of computational primitives as building blocks, the D3M system should synthesize a pipeline and set the appropriate hyper-parameters to solve a previously unknown data and problem. The D3M system also has a user interface that enables users to interact with and improve the automatically generated results \citep{blei2017datascience}.

\begin{figure}[h]
\begin{minipage}[b]{0.45\linewidth}
\centering
\includegraphics[width=0.9\linewidth]{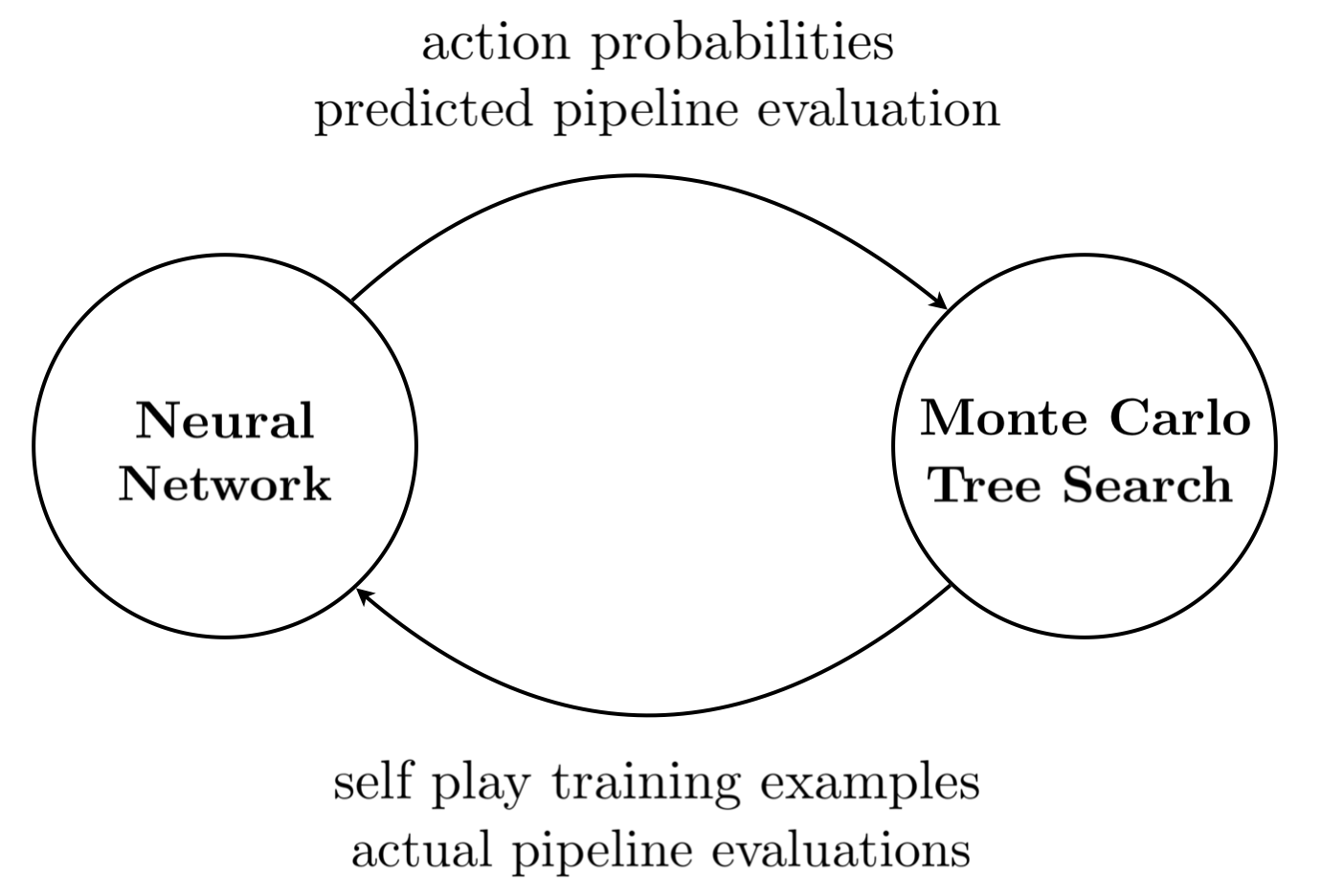}
\label{fig:alphazero}
\end{minipage}
\begin{minipage}[b]{0.45\linewidth}
\centering
\small
 \begin{tabular}{|l|l|l|} 
 \hline
  & AlphaZero & AlphaD3M \\
 \hline\hline
 Game & Go, chess & AutoML \\ 
 \hline
 Unit & piece & pipeline primitive \\
 \hline
 State & configuration & meta data, task, pipeline \\
 \hline
 Action & move & insert, delete, replace \\
 \hline
 Reward & win, lose, draw & pipeline performance \\
 \hline
\end{tabular}
\newline
\newline
\newline
\label{tab:game}
\end{minipage}
\caption{AlphaD3M iterative improvement (left); AlphaD3M game representation (right).}
\label{fig:alphazero-and-game}
\end{figure}

Inspired by AlphaZero~\citep{silver2017alpha0}, we frame the problem of pipeline synthesis for model discovery as a single-player game \citep{mcaleer2018solving}: the player iteratively builds a pipeline by selecting among a set of actions which are insertion, deletion, replacement of pipeline parts. An inherent advantage of this approach is that at the end of the process, once there is a working pipeline, it is completely explainable, including all the actions and decisions which led to its synthesis. Another advantage is that our approach leverages recent advances in deep reinforcement learning using self play, specifically expert iteration \citep{anthony2017thinking} and AlphaZero \citep{silver2017alpha0}, by using a neural network for predicting pipeline performance and action probabilities, along with a Monte-Carlo Tree Search (MCTS), as illustrated in Figure \ref{fig:alphazero-and-game} (left), which takes strong decisions based on the network. The process progresses by self play with iterative self improvement, and is known to be highly efficient at finding a solution to search problems in very high dimensional spaces.
We evaluate our approach using the OpenML dataset on the tasks of classification and regression, demonstrating competitive performance and computation times an order of magnitude faster than other AutoML systems.

Each of the existing AutoML systems uses any one of the following key elements individually: differentiable programming, tree search, evolutionary algorithms, and Bayesian optimization, to find the best machine learning pipelines for a given task and dataset. Differentiable programming, of which neural network backpropagation is a special case, is used for learning feature extraction and estimation \citep{ganin2014unsupervised} and for end-to-end learning of machine learning pipelines with differentiable primitives \citep{mitar2017}. Bayesian optimization methods are used for hyper-parameter tuning \citep{bergstra2012}. Both AutoWEKA \citep{autoweka2017} and Autosklearn \citep{feurer2015autosklearn} extend the application of these techniques to the selection of the model in addition to the hyper-parameter values, solving the combined algorithm selection and hyper-parameter optimization problem by fitting probabilistic models capturing the relationship between parameter values and performance measures using a Gaussian Process, Random Forest, or tree-structured Parzen estimator \citep{bergstra2011}. Auto-Tuned Models \citep{atm2017} represent the search space as a tree with nodes being algorithms or hyperparameters and searches for the best branch using a multi-armed bandit. TPOT \citep{olson2016tpot} and Autostacker \citep{chen2018autostacker} uses evolutionary algorithms to generate machine learning pipelines while optimizing their hyperparameters. TPOT represents machine learning pipelines as trees, whereas Autostacker represents them as stacked layers.

Our goal is to search within a large space for the machine learning, and pre and post processing primitives and parameters which together constitute a pipeline for solving a task on a given dataset. The problem is that of high dimensional search. Although the datasets differ, the solution pipelines contain recurring patterns. Just as a data scientist develops intuition and patterns about the pipeline components, we use a neural network along with a Monte-Carlo tree search in an iterative process. This combination results in the network learning these patterns while the search splits the problem into components and looks ahead for solutions. By self play and evaluations the network improves, incorporating a better intuition. An advantage of this iterative dual process is that it is computationally efficient in high dimensional search \citep{silver2017alpha0}.

\section{Methods}
Following dual process theory, we solve the meta learning problem by sequence modeling using a deep neural network and Monte Carlo tree search (MCTS) \citep{silver2017alpha0,anthony2017thinking}. This section describes our representation, followed by details of the neural network and MCTS.

\subsection{Representation}
Figure \ref{fig:alphazero-and-game} (right) illustrates a high level analogy between a two player competitive game and our single player pipeline synthesis game, including state, action, and reward. A pipeline is a data mining work flow, of pre-processing, feature extraction, feature selection, estimation, and post-processing primitives. Algorithm \ref{alg:pipelineencoding} describes our pipeline state representation. Our architecture models meta data and an entire pipeline chain as state rather than individual primitives. A pipeline, together with the meta data and problem definition is analogous to an entire game board configuration. The actions are transitions from one state (pipeline) to another.

\begin{algorithm}[htbp]

  {\caption{Pipeline Encoding}\label{alg:pipelineencoding}}%
{%
Given datasets $D$, tasks $T$, and a set of possible pipeline sequences $S_{1} ,\ldots,S_{n}$, from the available machine learning, and data pre and post processing primitives.
\begin{itemize}
  \item For each dataset $D_{i}$ and task $T_{j}$:
    \begin{enumerate}
        \item Encode dataset $D_{i}$ as meta data features $f(D_{i})$.
        \item Encode task $T_j$.
        \item Encode the current pipeline at time $t$ by a vector $S_t$.
        \item Encode action $f_a(S_t)$, so policy $\pi$ maps ($f(D_{i})$, $T_j$, $S_t$) to ${f_a(S_{1}),\ldots,f_a(S_{n})}$.
    \end{enumerate}
\end{itemize}
}%
\end{algorithm}

\subsection{Neural Network}
AlphaD3M uses a recurrent neural network, specifically an LSTM. Let $f_{\theta}(s)$ = $(P(s,a), v(s))$, where $P(s,a)$ is the action probabilities and $v(s)$ the evaluation score of the model predicted by the network $f$ with parameters $\theta$, for a given dataset $D$ and task $T$, for a given state $s$. The neural network predicts the probabilities over actions $a$ which lead to sequences $S$ that describe a pipeline, which in turn solves the given task on the dataset. The network inputs are training examples $(s_{t}, \pi_{t}, e_{t})$ from games of self play, where $s_{t}$ is the state at time $t$, $\pi_{t}$ the policy estimated by MCTS, and $e_{t}$ the actual pipeline evaluation at the end of the game. The state $s_{t}$ is composed of a vector encoded as described in Algorithm \ref{alg:pipelineencoding}. The network outputs are probabilities over actions $P(s,a)$, and an estimate of pipeline performance $v$.

We optimize the network parameters $\theta$ by making the predicted model $S$ match the real world model $R$ and the predicted evaluation results $v$ match the real world evaluation $e$, by minimizing the cross entropy loss between $S$ and $R$, and the mean squared error between $v$ and $e$. We add an $\ell_2$ regularization term for the network parameters $\theta$ to avoid over-fitting and an $\ell_1$ regularization term which prefers simple pipelines. Thus our network $f_{\theta}$ is trained by minimizing the following non-linear loss function using stochastic gradient descent:
\begin{equation}
\label{eq:lossfunction}
L(\theta) = S \log R + (v - e)^{2} + \alpha \| \theta \|_{2} + \beta \| S \|_{1}.
\end{equation}

\subsection{Monte Carlo Tree Search}
Our algorithm takes the predictions $(P(s,a), v(s))$ of the neural network and uses them in a MTCS by running multiple simulations to search for a pipeline sequence $R$ with a better evaluation. 
The search result $R$ improves upon the predicted result $S$ given by the network by improving the network policy using the update rule:
\begin{equation}
    U(s,a) = Q(s,a) + c P(s,a) \frac{\sqrt{N(s)}}{1 + N(s,a)},
\end{equation}
where $Q(s,a)$ is the expected reward for action $a$ from state $s$, $N(s,a)$ is the number of times action $a$ was taken from state $s$, $N(s)$ the number of times state $s$ was visited, $P(s,a)$ is the estimate of the neural network for the probability of taking action $a$ from state $s$, and $c$ is a constant which determines the amount of exploration. At each step of the simulation, we find the action $a$ and state $s$ which maximize $U(s,a)$ and add the new state to the tree if it does not exist with the neural network estimates $(P(s,a),v(s))$ or call the search recursively otherwise. Next, the model represented by $R$ is realized and applied to the data to solve the task, resulting in a better evaluation $e$ which is the result of running the generated pipeline $R$ on the data and task. Thus the real world search provides us with $(R, e)$, where $R$ is the real world model, consisting of machine learning primitives, and $e$ the real world evaluation of the model and pipeline using those primitives on the data and task.

The neural network predictions, the MCTS model, and the real world evaluation, together, define a loss function shown in Equation \ref{eq:lossfunction}, which is minimized to improve the neural network parameters. This process continues iteratively until the best model, which automatically solves the task, is found. 

Inspired by the neural editor \citep{guu2017generating} we use edit operations that make the pipeline generation explainable by design. For each iteration of self play the MCTS searches the possible valid pipelines. For each state or pipeline the next possible states or pipelines are limited to those derived from the edit operations of the current state.

\section{Results}
The data consists of 313 different tabular datasets, of which 296 are from OpenML \citep{openml2014}. We considered classification, both binary (121 datasets) and multi-class (108 datasets), and univariate regression tasks (84 datasets). Baseline pipelines were constructed using sklearn SGD estimators for classification and regression, and an annotated tabular feature extractor which uses linear SVC, Lasso, percentile classification or regression estimators from sklearn.

\begin{figure}
\centering
\includegraphics[width=0.4\linewidth]{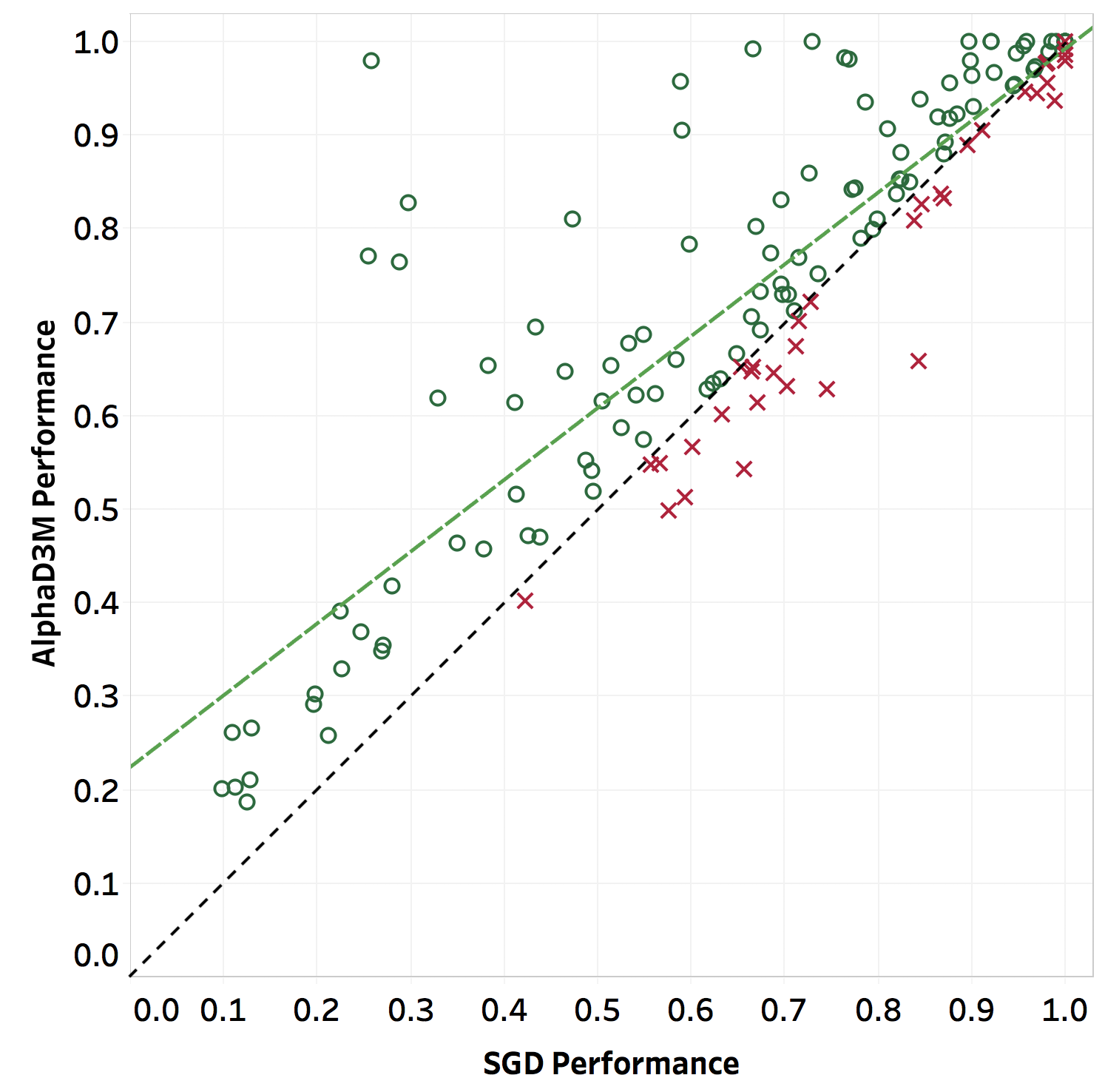}
\caption{AlphaD3M vs. SGD performance for 180 classification tasks on OpenML datasets.}
\label{fig:alphad3m_sgd_cv}
\end{figure}

\begin{figure*}
\centering
\includegraphics[width=1\linewidth]{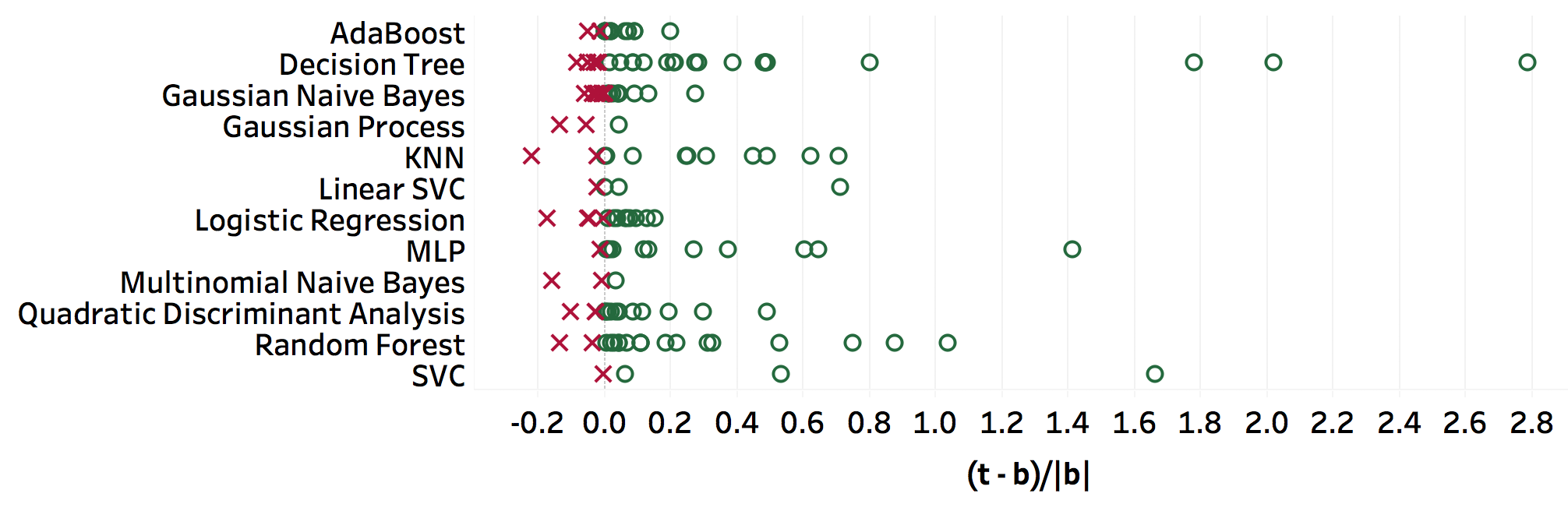}
\caption{Comparison of normalized AlphaD3M performance $t$ with SGD baseline performance $b$ by estimator.}
\label{fig:primitives_usage}
\end{figure*}

Figure \ref{fig:alphad3m_sgd_cv} compares performance between AlphaD3M and SGD which is the baseline pipeline. Each of the 180 points represents a classification task on a different OpenML dataset. The datasets for which AlphaD3M performs better than SGD are shown by green circles and those for which SGD performs better are shown by red crosses. Figure \ref{fig:alphad3m_sgd_cv} shows that AlphaD3M performs better than baseline for 75\% of the datasets, both are comparable for 18\% of the datasets, and performs worse for only 7\% of the datasets. Figure \ref{fig:primitives_usage} shows the normalized difference in cross validation performance of AlphaD3M $t$ and SGD baseline $b$ for a classification task for 180 datasets, split according to the estimators used by AlphaD3M, demonstrating better performance across diverse estimators.

\begin{figure*}
\centering
\includegraphics[width=1\linewidth]{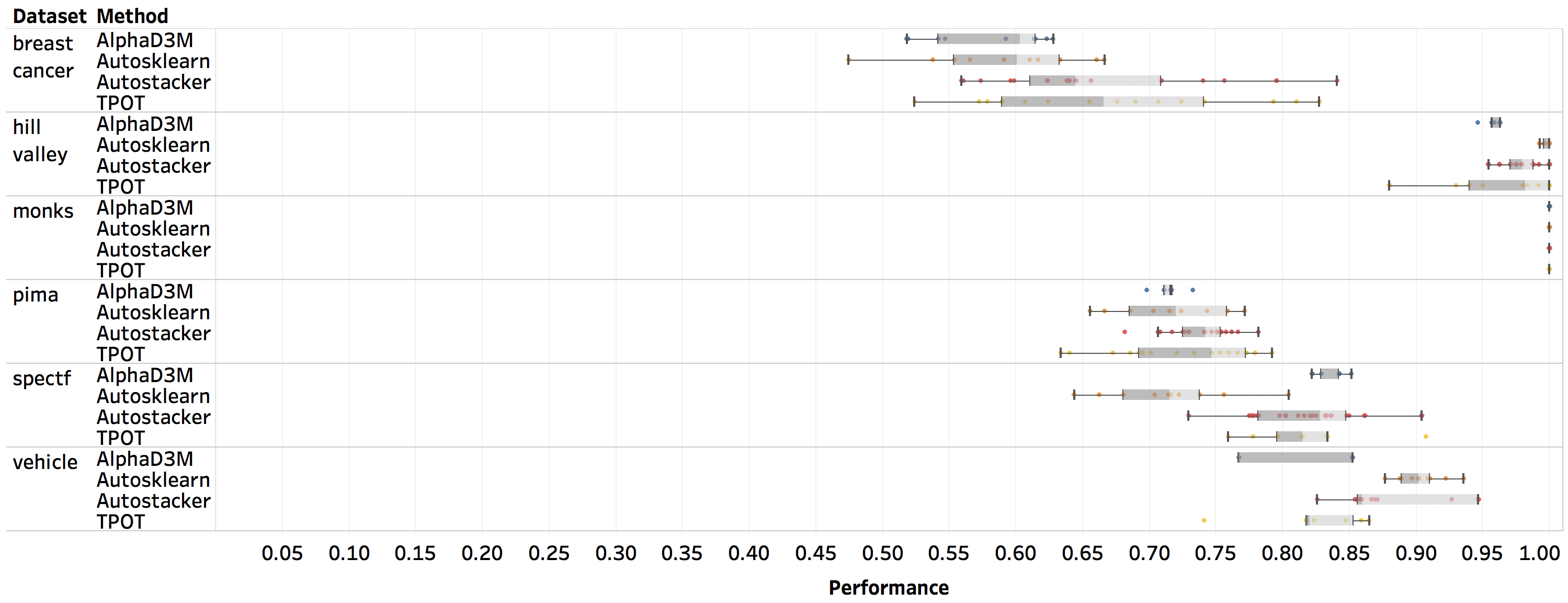}
\caption{Comparing between performance of AutoML methods on OpenML datasets.}
\label{fig:performancecomparison}
\end{figure*}

Figure \ref{fig:performancecomparison} compares performance between different AutoML methods: Autosklearn, TPOT, and Autostacker, and our method AlphaD3M, for a number of common OpenML datasets, which serve as representative benchmark datasets for AutoML systems \citep{olson2017pmlb,olson2016tpot,chen2018autostacker}. For each method and dataset, we compute the performance mean and standard deviation by repeated evaluation. As shown in Figure \ref{fig:performancecomparison} our method, AlphaD3M, is competitive with other approaches. All four methods are competitive and on par, as their performance including confidence intervals intersect; whereas SGD and Random Forest are not competitive with the leading AutoML methods.

AlphaD3M is implemented using PyTorch. Our implementation takes advantage of GPUs while training the neural network and uses CPUs for the MCTS. Table \ref{tab:time} compares the running time of TPOT, Autostacker, and AlphaD3M on the same datasets, along with the corresponding speedup factors. Table \ref{tab:time} shows that AlphaD3M performs on average an order of magnitude faster, reducing computation time from hours to minutes.

\begin{table}[]
\centering
\small
\caption{Running time comparison (in seconds and speedup factors).}
\label{tab:time}
\begin{tabular}{lllllll}
\hline
Dataset/Method & TPOT & Autostacker & AlphaD3M & Speedup vs TPOT & Speedup vs AS\\
\hline
breast cancer & 3366 & 1883 & 460 & 7.3 & 4\\
hill valley & 17951 & 8411 & 556 & 32.2 & 15.1\\
monks & 1517 & 1532 & 348 & 4.3 & 4.3\\
pima & 5305 & 1940 & 619 & 8.5 & 3.1\\
spectf & 4191 & 1673 & 522 & 8 & 3.2\\
vehicle & 16795 & 4010 & 531 & 31.6 & 7.5\\
\end{tabular}
\end{table}

\section{Conclusions}
We introduced AlphaD3M, an automatic machine learning system with competitive performance, which is an order of magnitude faster than existing state-of-the-art AutoML methods, reducing computation time from hours to minutes. We presented the first single player AlphaZero game representation applied to meta learning by modeling meta-data, task, and entire pipelines as state.

\bibliography{paper}

\end{document}